# InsightFlow: LLM-Driven Synthesis of Patient Narratives for Mental Health into Causal Models

Shreya Gupta[1], Prottay Kumar Adhikary[2], Bhavyaa Dave[2], Salam Michael Singh[3], Aniket Deroy[2], Tanmoy Chakraborty[2]*

[1] Department of Mathematics, IIT Delhi, Delhi, India
[2] Department of Electrical Engineering, IIT Delhi, Delhi, India
[3] Department of Computer Science and Engineering, IIIT Manipur, India

* Corresponding author: tanchak@iitd.ac.in

## Abstract

**Background:** Clinical case formulation in mental health involves structuring patient symptoms, history, and psychosocial factors into a causal model. This often uses the "5P" framework of presenting problems, predisposing factors, precipitating factors, perpetuating factors, and protective factors. Manually creating these causal graphs from therapy transcripts is time-consuming and often varies between experts. Large language models (LLMs) offer a way to automate the extraction of such information from text.

**Objective:** We developed InsightFlow, an LLM-based method to automatically generate 5P-aligned causal graphs from patient-therapist dialogues. We compared its output against clinician-annotated graphs.

**Methods:** We collected 46 psychotherapy intake session transcripts and had clinical experts manually annotate causal factors using a structured protocol. We then prompted Llama-3 8B to identify relevant factors and iteratively refine connections between them. We compared the LLM-generated graphs to human-generated graphs using automated graph similarity measures (NetSimile, edge and node embedding similarities) and a clinical rubric. Expert reviewers rated completeness, consistency, specificity, plausibility (of nodes and edges), and utility on a 5-point scale. We also computed graph metrics (density, centrality, clustering, distance, and communities) to compare structures.

**Results:** The LLM-generated graphs had a structural similarity (NetSimile) in the range of 0.38-0.46 when compared to each human annotator's graph. Inter-annotator similarity was about 0.45. Semantic similarity of edges and nodes was high (mean cosine > 0.70), indicating the models agreed on most core concepts. Experts gave the LLM graphs moderate scores (mean around 3.2-3.6 out of 5) in all evaluation categories, similar to the range for human graphs.

Quantitatively, LLM graphs had slightly higher density (0.27 vs ~0.25 in human graphs) and clustering (local clustering ~0.17 vs ~0.12) but similar diameter and path length. Centrality metrics showed LLM graphs distributed influence more evenly (higher betweenness and closeness). Community analysis revealed that nodes in the LLM graph are grouped logically by type. Overall, LLM outputs matched human graphs in complexity and covered similar content, though human graphs tended to form chain-like paths while LLM graphs formed richer, web-like networks.

**Conclusions:** Our findings demonstrate that a structured LLM approach can generate clinically meaningful case formulation graphs that are comparable to expert work. While stylistic differences exist (LLM webs vs human chains), the automated outputs fall within the natural variability of human formulations. *InsightFlow* suggests a promising tool to augment therapists by automating causal modeling of patient narratives. Future work should refine temporal reasoning and prune redundancies to further match expert reasoning. Future work should address these issues and explore larger datasets.

**Keywords**

large language models; 5P framework; causal graph; case formulation; clinical narrative; mental health; graph similarity

**Introduction**

Clinical case formulation is a systematic way to explain a patient's problems by listing symptoms, contextual factors, and maintaining mechanisms [1, 2]. Psychotherapists often draw *causal graphs* where each node is a psychological factor (symptom, trigger, etc.) and edges represent hypothetical cause-and-effect relations [3]. These visual models help in understanding and treating a patient by revealing how issues are interconnected. However, building such graphs manually from therapy transcripts is time-intensive and subjective [4, 5]. Even experienced clinicians can disagree on what factors are most important or how they are linked; for example, inter-rater agreement (Cohen's kappa) is often low (around 0.33-0.45) among trained raters constructing case formulations [6].

Large language models (LLMs) like GPT and Llama-3 have shown promise in reading and structuring text [7]. They can recognize entities and organize unstructured information into structured outputs. Recent work suggests LLMs might replicate human reasoning patterns when asked to generate explanations or chains of thought [8, 9]. This raises the possibility of automating case formulation by prompting an LLM to read therapy transcripts and output causal graphs [10, 11, 12]. The theoretical foundation for this approach is grounded in the literature on structural causal models (SCMs) and directed acyclic graphs (DAGs), which provide a rigorous mathematical framework for representing and reasoning about causal relationships and interventions [13]. In mental health, clinicians use frameworks like the 5P model to structure case formulations. This model identifies **Predisposing** factors (long-term vulnerabilities), **Precipitating** events (recent triggers), **Perpetuating** patterns (behaviors or conditions that maintain problems), **Presenting** problems (current symptoms), and **Protective** factors (strengths) [14]. By implementing a structured approach to otherwise scattered patient narratives, this framework helps therapists systematically identify vulnerabilities, triggers, and maintaining factors. Evidence shows that case formulation enhances treatment planning, improves therapeutic alliance, and increases teamwork [15].

Our goal is to create an end-to-end system that takes a therapy conversation and produces a 5P-based causal graph. This involves prompt engineering and validation of the generated graph against human experts [16]. To our knowledge, this is among the first attempts to directly use LLMs for building clinical case formulation graphs from dialogue. Applications include helping therapists quickly outline formulations, monitoring therapy progress, and analyzing large sets of anonymized sessions for research or public health [17, 18]. Our contributions are: (1) Developing a benchmark by manually annotating 46 intake sessions using a 5P-based protocol; (2) Building a novel iterative LLM prompting method with feedback to identify factors and causal links; (3) Comparing this automated approach against a human-built graph using both computational metrics and expert clinical ratings; (4) Analyzing structural differences via graph-theory measures and community detection; and (5) Documenting limitations (e.g. redundant nodes, lack of temporal flow) and suggesting improvements. The results show the LLM-generated graphs are within the range of human expert variability and capture the key elements needed for case formulation.

## Methods

This study followed a structured approach to develop, implement, and evaluate an automated system for generating causal graphs from therapy conversations. The overall methodology involved dataset preparation, manual annotation to create ground truth graphs, automatic graph generation using a large language model, and both computational and expert-based evaluation. The aim was to ensure that the generated graphs are clinically meaningful, structurally comparable to human formulations, and useful for real-world applications.

### Study Design

We used a mixed-method design combining computational evaluation and human expert assessment. The study consisted of three main stages. First, a benchmark dataset of therapy conversations was annotated manually using a structured protocol based on the 5P framework. Second, an LLM-based system was used to automatically generate causal graphs from the same conversations. Third, the generated graphs were compared with human annotations using both graph similarity metrics and expert evaluation.

**Dataset**

We used 46 clinical intake session transcripts from the MentalCLOUDS [19] corpus, which contains therapist–patient dialogues. These conversations covered a range of presenting issues (e.g., anxiety, addiction, life transitions) and contexts. To control for annotation, we split the data into two sets of 23 sessions each. Two trained annotators independently reviewed each set. On average, each session contained 1,837 words (overall), with Group 1 sessions slightly shorter (1,736 words) than Group 2 (1,937 words). Sentence counts averaged about 203 per session. The number of turns (speaker changes) varied: therapists spoke ~110 turns per session on average overall, patients about 93 turns. The average length of the utterance was 9.5 words. These numbers indicate the dialogues were roughly similar in size across groups. Having two annotators per session and a variety of cases ensured that our evaluation covered both typical and diverse clinical scenarios.

Since no existing gold-standard dataset of mental health causal graphs was available, we created one. We used the 5P framework to guide annotation. Protective factors were sparse in intake sessions, so we focused on the other four categories. Two clinicians read each transcript and followed a step-by-step procedure to build a causal graph:

1. **Identify Presenting Problems:** Extract the main symptoms or issues mentioned by the patient and therapist (e.g., "loss of appetite", "insomnia"). These form initial nodes.
2. **Find Precipitating Factors:** Look for recent events or stressors tied to the onset of the problems (e.g. "failing an exam", "job loss").
3. **Trace Predisposing Factors:** Determine underlying vulnerabilities that made the patient susceptible (e.g. "family history of anxiety", "years of unemployment").
4. **Add Perpetuating Factors:** Include any ongoing behaviors or conditions that keep problems going (e.g. "rumination", "social isolation").
5. **Connect the Factors:** Link nodes with directed edges to form causal chains. We started from the presenting problem nodes and traced backwards to causes and forwards to effects. For example, if lack of sleep were a symptom, we might connect "noise pollution" (precipitating) to "sleep disturbance" (presenting) to "daytime fatigue" (secondary effect). Each potential edge was added only if the transcript context supported that cause-effect link.
6. **Iterate and Refine:** We repeated this mapping recursively, expanding deeper layers of causes until no new plausible factors were evident. Annotators checked for redundant or spurious links and removed them.

Figure 1 illustrates the iterative construction process across four stages. By the end, each transcript had a directed graph of nodes labeled as "Presenting", "Predisposing", "Precipitating", or "Perpetuating". These manually crafted graphs constitute our ground truth against which we compare the automated output.

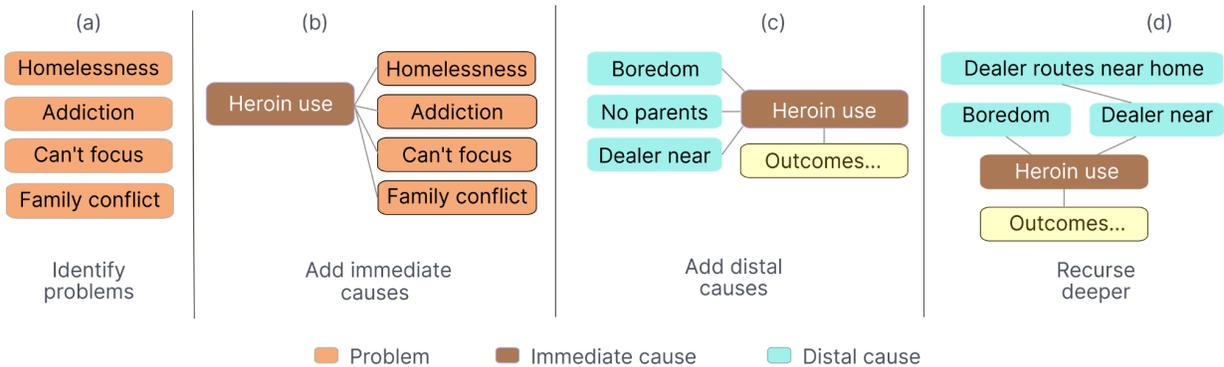

Figure 1: Iterative construction of a clinical causal graph using the 5P framework, illustrated with an example addiction case across four stages. (a) Annotation begins by identifying all presenting problems reported in the intake session (e.g., risk of homelessness, addiction, difficulty concentrating, and family conflict), which are placed as isolated nodes with no edges. (b) Immediate causes are introduced: repeated heroin use is identified as the proximal cause and linked via directed edges to the presenting problems, forming the first causal layer. (c) Distal causes are added one level further back: boredom, parental absence, and proximity to a drug dealer are each connected to heroin use, extending the graph upward. (d) The process recurses: the dealer's proximity is itself explained by a contextual factor (the dealer's routine route passing near the patient's home), adding a fourth causal layer. This recursive backward-chaining continues until no further plausible causes can be identified from the transcript.

**Automatic Graph Generation**

Our automated system, InsightFlow, used Llama-3 8B to build the graph in two stages. First, we prompted the model to extract factor nodes from the conversation. A tailored prompt asked the LLM (framed as a clinical psychologist) to list in JSON format the presenting problems, predisposing factors, precipitating factors, and perpetuating factors that apply to the dialogue. For example, we asked: *"Identify the patient's presenting problems, predisposing factors, precipitating factors, and perpetuating factors from the conversation below."* The model returned lists of phrases for each category. We parsed these into graph nodes.

Second, we determined which pairs of nodes should be connected by a causal edge. We considered all potential directed pairs across categories (e.g., a precipitating node might cause a presenting node). For each pair, we created a new prompt asking if the first node causes the second, providing the relevant text for context. The LLM responded with "TRUE" or "FALSE" (formatted in JSON). If it indicated "TRUE", we added a directed edge between those nodes in the graph. This pairwise check used a second "feedback" Llama prompt that had access to the full conversation to judge causality. In this way, the model filtered out unlikely links and built a coherent causal graph. Figure 2 illustrates the complete InsightFlow pipeline, from raw transcript input through LLM-based factor extraction to final graph construction.

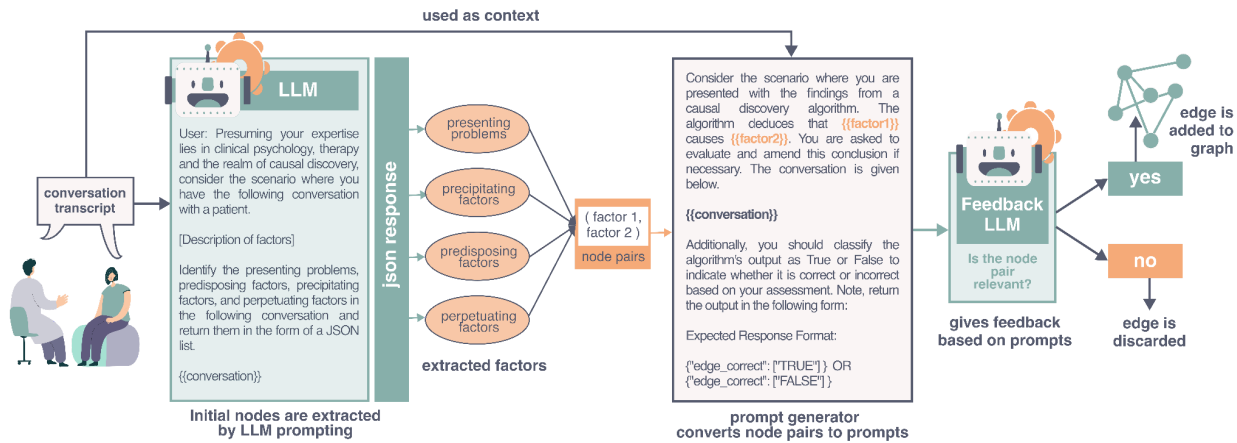

*Figure 2: Complete pipeline of the InsightFlow system for constructing causal graphs from psychotherapy conversations. The process begins with a therapy conversation transcript, which is provided as input to a large language model (LLM) prompted with domain-specific instructions grounded in clinical psychology. In the first stage, the LLM extracts structured factors corresponding to the 5P framework, including presenting problems, predisposing factors, precipitating factors, and perpetuating factors. These extracted elements are returned in a structured JSON format and form the nodes of the graph.*

This feedback-based approach encouraged the LLM to reason about each relationship explicitly. It also helped avoid repeating edges: we kept a running list of identified edges, and node extraction was done in one go to avoid repetition. The result was a fully formed causal network with nodes from the 5P categories and edges representing inferred cause-and-effect relations. We used an NVIDIA A100 GPU with 80 GB of memory for the generation.

**Evaluation**

**Automated Metrics**

To quantify similarity between each LLM-generated graph and its human reference, we used both structural and semantic measures. Structural similarity was measured by NetSimile [20], which compares summary features of the graphs (degree distribution, clustering, etc.). This yielded a score between 0 (completely different) and 1 (identical structure). For semantic similarity, we embedded node-pair labels using SBERT [21] and computed cosine similarity between matched edges across the two graphs, averaging the best-match scores to estimate relational alignment independent of surface wording. Node-level similarity was assessed both globally (by averaging node embeddings) and by degree, comparing the most structurally prominent nodes across graphs. To further illustrate how a therapy conversation maps onto a causal graph in practice, Figure 3 presents a worked example alongside the dialogue that produced it.

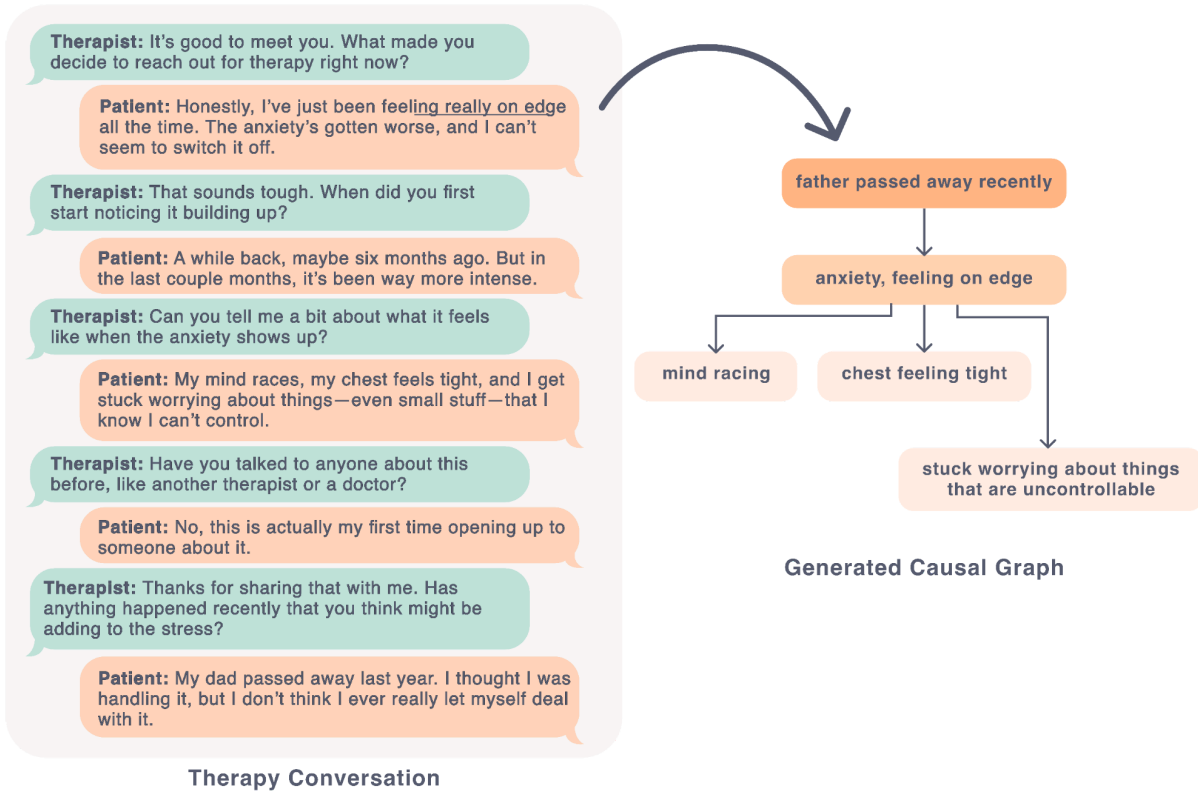

Figure 3: An illustrative example of how a therapy conversation is transformed into a causal graph using the proposed framework. The left panel presents a segment of a therapist–patient dialogue, where the patient describes persistent anxiety, physiological symptoms (e.g., chest tightness), and cognitive patterns (e.g., uncontrollable worrying), along with a recent life event (bereavement). The right panel shows the corresponding causal graph generated by the system.

**Human Expert Review**

Five clinical experts, advanced psychology trainees and practitioners aged 22–35, with 6 months to 4 years of experience, independently evaluated each LLM-generated graph blind to its origin across six dimensions on a 5-point scale (1 = very poor, 5 = excellent): Completeness (whether all relevant problems and their causal layers were captured), Consistency (alignment of causal relations with psychological literature), Specificity (whether nodes were pitched at an appropriate level of clinical granularity, neither too broad nor too narrow), Plausibility of Nodes (proportion of nodes traceable to explicit transcript statements), Plausibility of Edges

(proportion of directed links supported by dialogue evidence), and Utility (whether the graph could realistically inform clinical decision-making or treatment planning without substantial manual refinement).

Raters were trained on detailed rubric definitions before evaluation. For each dimension, scores were anchored at both ends: for example, a Completeness score of 1 indicated most problems were absent and the graph lacked depth, while a 5 required full coverage of presenting problems, triggers, and deep predisposing factors; a Utility score of 1 meant the graph offered no actionable insight, while a 5 indicated it could directly support clinical practice with little modification. Mean scores were computed per metric, and inter-rater agreement was assessed using Fleiss' kappa.

**Additional Graph Analyses**

To explore structural patterns, we computed graph theory features for each graph (LLM and human). We measured edge density (fraction of all possible edges that are present), as well as centrality measures (degree [22], betweenness [23], closeness [22] for each node). We averaged these across nodes to compare overall connectedness. We also calculated clustering coefficients (local and global) and counted simple motifs (triangles) [25, 26, 27]. To see how factors were grouped, we ran three community detection algorithms – Leiden [28], Girvan-Newman [29], and Infomap [30], on each graph. We analyzed whether communities contained nodes of the same 5P category or related categories. Finally, we computed distance metrics (Kullback-Leibler divergence [31] and Earth Mover's Distance [32]) between the degree distributions of graphs to quantify global differences.

**Results**

Across all 46 sessions, the LLM-generated graphs were broadly comparable in size to the human-annotated graphs, producing a similar number of nodes and edges on average. Overall similarity scores, structural analyses, and expert ratings converged on the same finding: the automated output falls within the range of human variability, though with a consistently denser and more interconnected style. We report these findings across annotation agreement, automatic similarity, graph topology, and expert evaluation in turn.

Inter-annotator agreement on graph content was modest, consistent with the known subjectivity of clinical case formulation. Fleiss' kappa across the six evaluation dimensions ranged from approximately -0.02 to 0.07, indicating slight to negligible reliability, a pattern that mirrors prior findings on variability in clinical judgment. For context, one annotator group reached κ = -0.02 for node plausibility and κ = 0.07 for consistency. Structural similarity between the two human annotators' own graphs averaged NetSimile = 0.45 overall and 0.53 in Group 2, further reflecting how differently trained clinicians can formulate the same case. When the three experts rated the LLM graphs, total scores (out of 30) were 22.3, 19.3, and 19.7 (mean ≈ 20.4, SD ≈ 4.1), suggesting moderate but not uniform agreement on quality.

| Group | Comparison | NetSimile | | Mean Edge | | Node Set | | Node Centrality | |
|---|---|---|---|---|---|---|---|---|---|
| | | Mean | Std. | Mean | Std. | Mean | Std. | Mean | Std. |
| GROUP 1 | A vs B | 0.3735 | 0.0906 | 0.707 | 0.2374 | 0.7758 | 0.1141 | 0.2205 | 0.0842 |
| | Auto vs A | 0.3996 | 0.0976 | 0.7123 | 0.168 | 0.7853 | 0.0465 | 0.2024 | 0.0955 |
| | Auto vs B | 0.4604 | 0.1243 | 0.643 | 0.2651 | 0.7421 | 0.0834 | 0.1902 | 0.0712 |
| GROUP 2 | A vs B | 0.5254 | 0.1433 | 0.7327 | 0.1724 | 0.844 | 0.0649 | 0.2037 | 0.0789 |
| | Auto vs A | 0.3605 | 0.0799 | 0.6963 | 0.1678 | 0.7316 | 0.0905 | 0.1443 | 0.0668 |
| | Auto vs B | 0.3955 | 0.1145 | 0.6747 | 0.1579 | 0.7268 | 0.0943 | 0.1256 | 0.0503 |
| TOTAL | A vs B | 0.4495 | 0.1412 | 0.7199 | 0.2054 | 0.8099 | 0.098 | 0.2121 | 0.0811 |
| | Auto vs A | 0.38 | 0.0903 | 0.7043 | 0.1661 | 0.7584 | 0.0761 | 0.1733 | 0.0866 |
| | Auto vs B | 0.428 | 0.1226 | 0.6589 | 0.2163 | 0.7344 | 0.0883 | 0.1579 | 0.0691 |

*Table 2: Automatic evaluation results comparing human-annotated and LLM-generated causal graphs using multiple similarity metrics (NetSimile, Mean Edge Similarity, Node Set Similarity, and Node Centrality Similarity) across two groups of annotators.*

Turning to automatic similarity (Table 2), LLM graphs achieved NetSimile scores of 0.38-0.46 against human graphs, falling within the 0.37-0.53 range observed between the two human annotators themselves. Notably, in Group 1, the LLM achieved slightly higher similarity to individual annotators (0.40 and 0.46) than the annotators achieved with each other (0.37), suggesting the automated output is no more divergent from a given clinician than that clinician's colleague. Semantic overlap was consistently strong: mean edge similarity (cosine similarity of SBERT-embedded edge pairs) was approximately 0.70, and node-set similarity averaged 0.75–0.80 across both groups. The relatively low standard deviations for node-set similarity indicate that high factor overlap was maintained reliably across sessions, not just on average. Standard deviations for edge similarity (0.15-0.27) indicate that semantic correspondence remained above 0.5 for the majority of graphs. Taken together, these figures indicate that the LLM identified largely the same clinical factors as the clinicians, even when surface wording differed.

Examining topological properties (Figure 4a), LLM graphs were consistently denser than human graphs, with a mean edge density of 0.29 versus 0.25, meaning the model included more connections per graph. This density difference was stable across both groups and is visible in the comparative bar plots. A closer look at clustering (Figure 4b) reveals that LLM graphs formed more triangular motifs: an average of 1.06 triangles per graph compared to 0.40 and 0.14 for the two human annotators. Global transitivity followed the same direction, with LLM graphs at

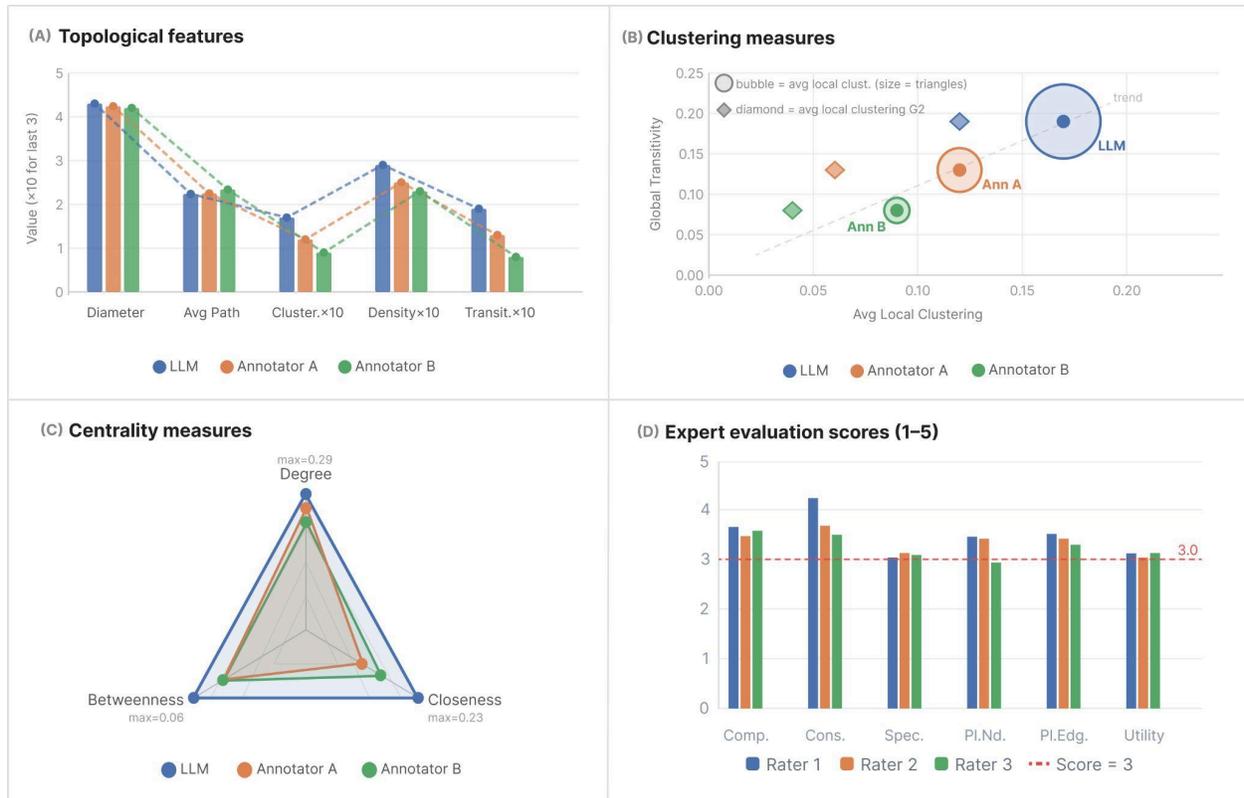

*Figure 4: Comparative analysis of LLM-generated and human-annotated causal graphs across structural and expert evaluation metrics. InsightFlow (LLM, blue) is compared with Annotator A (orange) and Annotator B (green). **(a) Topology:** graphs are similar in diameter (~4.2-4.3) and path length (~2.2-2.3), with LLM showing slightly higher edge density (0.29 vs ~0.25) and transitivity. **(b) Clustering:** LLM graphs show higher triangle counts (1.06 vs 0.40 and 0.14) and global transitivity (~0.19 vs ~0.13 and ~0.08), indicating more triadic closure. **(c) Centrality:** degree is comparable (~0.29 vs ~0.26 and ~0.23), while LLM has higher betweenness and closeness, suggesting more distributed influence. **(d) Expert ratings:** mean scores across six 1–5 dimensions from three blinded raters; most values fall between 3.0 and 4.3, with Consistency highest and Specificity and Utility lowest (red dashed line = score 3 threshold).*

~0.19 versus ~0.13 and ~0.08 for human annotators. Where a clinician might construct a lean causal chain A→B→C, the LLM frequently added the shortcut A→C, producing triadic closure. This reflects the model's exhaustive pairwise evaluation strategy rather than clinical error; the extra links were generally regarded as plausible by experts. Human annotators, by contrast, produced strongly disassortative graphs with chain-like formulations focused on a smaller set of core clinical problems, which aligns with how clinicians typically prioritize parsimony in case conceptualization. Variability in the LLM-generated graphs was also lower than in human-constructed graphs, suggesting the model behaves as a stable, internally consistent annotator. Differences between conversation groups were more pronounced than differences between generation methods, indicating that session content drives graph topology more than annotator identity.

Centrality patterns (Figure 4c) reinforce this further. Average degree centrality was broadly comparable across sources (~0.29 for LLM vs. ~0.26 and ~0.23 for the two human annotators), but LLM graphs showed higher betweenness and closeness centrality distributed across more nodes, meaning causal influence was spread more evenly across the graph. Human annotators concentrated causal flow through one or two focal nodes, a hub-and-spoke structure that reflects deliberate clinical prioritization. Community detection results were consistent with the topological findings: the Leiden algorithm found, on average, 3.55 communities in LLM graphs compared to 3.22 and 3.98 for the two human annotator sets. The Girvan-Newman method produced a similar pattern, Infomap detected slightly more communities in LLM graphs, and label propagation showed greater variance but comparable central tendencies. Crucially, these communities generally aligned with 5P categories. In a representative addiction case, for example, one community grouped the presenting factor ("struggling with heroin addiction") with directly reinforcing perpetuating factors, a second grouped precipitating environmental factors with a perpetuating consequence, and a third captured indirect perpetuating factors such as lack of alternative coping mechanisms, a structure that mirrors the kind of thematic clustering a clinician would recognize as clinically coherent. The inter-human divergence on distributional distance measures was also substantial (mean KL divergence = 4.10, Earth Mover's Distance = 0.61), underscoring baseline subjectivity; LLM-to-human distances were higher but remained within the same order of magnitude (KL = 5.23-5.57, EMD = 1.20-1.42).

Analysis of distances between 5P factor categories revealed further structural patterns. In human-generated graphs, the strongest connectivity was between presenting and perpetuating factors, reflecting a clinical emphasis on maintaining mechanisms. LLM-generated graphs showed relatively stronger connections between presenting and precipitating factors, suggesting a greater emphasis on proximal triggers. Predisposing factors showed a moderate association with perpetuating factors in both graph types, consistent with the theoretical claim that underlying vulnerabilities and maintaining conditions are conceptually related. The overall inter-category correlations were weak, supporting the conceptual separability of the 5P domains within the constructed graphs.

Expert evaluation scores were moderately positive across all dimensions (Figure 4d). Mean ratings on the 1-5 scale were approximately 3.3 for Completeness, 3.4 for Plausibility of Nodes, 3.4 for Plausibility of Edges, and 3.2-3.3 for Consistency, Specificity, and Utility, where a score of 3 denotes acceptable quality requiring some refinement. Group 2 graphs were judged to be more clinically coherent, detailed, and realistic, with causal links that better reflected real clinical reasoning; Group 1 graphs, while slightly lower in rated quality, were considered more practical and easier to use, likely owing to their relative simplicity. Inter-rater agreement on these ratings was low (Fleiss' κ near zero for most metrics), again reflecting the inherent subjectivity of the task. Nevertheless, the aggregate picture, expert totals averaging 20.4 out of 30, structural similarity within human variability bounds, and semantically grounded nodes traceable directly to transcript statements, indicates that InsightFlow's formulations are reasonably useful as a decision-support starting point, even if they do not yet fully substitute for expert clinical judgment.

## Discussion

### Principal Findings

Our study shows that it is feasible to use an LLM to automate the creation of 5P-based case formulation graphs from therapy dialogues. The LLM (Llama-3 8B) was able to read patient-therapist conversations and identify psychological issues and factors almost as well as trained clinicians. When comparing graphs, the similarity metrics and expert scores suggest that the LLM's output falls within the range of human variability. In other words, the LLM performed comparably to one clinician when judged by another.

The LLM graphs had advantages in certain areas. Because it exhaustively evaluated all pairs of factors, the LLM often uncovered connections that humans missed. This led to richer graphs (higher density and clustering). Experts saw these extra links as plausible, not simply noise. Moreover, since the LLM tended to reuse wording from the transcript, each node can be easily traced back to the source conversation. This traceability is valuable for transparency: a therapist reviewing the graph can quickly find the patient statement supporting each factor.

However, the LLM also had limitations. The graphs lacked any explicit time ordering. Factors were presented as a flat network of causes without indicating which happened first. Human clinicians often use the sequence of events to decide causality. Also, the LLM sometimes created redundant nodes (e.g., "increased stress" and "stress increasing") and broad factors that were split into pieces. This inflated the graph without adding new information. These are areas for improvement.

**Comparison with Prior Work**

Automated causal modeling in psychotherapy is a new area. Our approach is related to research on LLMs and chain-of-thought prompting in complex reasoning tasks. Prior studies have shown that LLMs can do causal reasoning and structured QA, but applying them to therapy case formulation is novel. We found, consistent with general NLP findings, that explicit multi-step prompts (node extraction then verification) improved performance over a single-shot prompt. Unlike purely statistical methods, our system embeds domain knowledge (5P definitions) into the prompts, which helps focus the LLM on clinically relevant aspects.

In comparison to other AI tools in mental health, which often focus on symptom detection or chatbot interactions, InsightFlow addresses a different need: the structured synthesis of rich narrative data. By building graphical models of patient cases, this tool could complement clinical work by providing a clear, visualized summary of factors. It is comparable to co-created therapy aids or case formulation tools, but is driven by AI rather than relying on manual input.

**Limitations and Future Work**

This study has limitations. The dataset of 46 sessions, while real, is modest in size and scope. We did not include diverse cultural or clinical populations, which could affect factor prevalence. The LLM used is relatively small (8B parameters); larger models or those fine-tuned on clinical text might do even better. Our evaluation also has subjectivity. The expert ratings showed low agreement, so the numeric averages must be interpreted cautiously. We aimed for a realistic clinical context, but further work could involve more clinicians or even field testing in actual therapy settings.

Technically, future work should improve temporal understanding and redundancy removal. For example, using a timeline-aware model or additional prompting could help the LLM order events correctly. Post-processing steps could merge near-duplicate nodes. We could also incorporate feedback from clinicians in the loop to refine the graph.

**Conclusions**

InsightFlow demonstrates that LLMs can effectively be harnessed to produce case formulation graphs from patient dialogues. The generated models are largely consistent with what experts would produce and show promise as a decision-support tool. They can potentially save clinicians time and highlight relationships that may not be immediately apparent. As LLMs continue to improve and are adapted to healthcare contexts, such automated formulation tools may become increasingly accurate and valuable. Careful integration, with attention to ethical use and expert oversight, will be important for real-world adoption.

## Acknowledgments

We thank Minoti Nilesh Karnik (M.A. Counselling Psychology), Avani Kane (M.Sc. Lifespan Counselling), Saniya Pathare (M.A. Lifespan Counselling), and Reena Rawat (M.Phil. Clinical Psychology) for their contributions to data annotation and evaluation.

## Funding

TC would like to thank the financial support of the Tower Research Capital Markets toward using machine learning for social good, and the Rajiv Khemani Young Faculty Chair Professorship in AI.

## Conflicts of Interest

The authors report no conflicts of interest.

## Data Availability

All data generated or analyzed during this study are included in this published article. The code and data are available at: https://github.com/ai4mhx/InsightFlow

## Authors' Contributions

SG conceptualized the study, designed the methodology, implemented the software, performed the formal analysis, curated the data, and wrote the original draft. PKA contributed to the initial ideation of the study and participated in the final drafting and refinement of the manuscript. BD provided domain expertise, handled annotation and evaluation framework design, conducted qualitative analysis and validation, and co-wrote the original draft. SMS contributed to methodology and theoretical guidance, and performed visualization guidance and editing. AD developed the evaluation design, conducted statistical analysis and validation, and reviewed the writing. TC contributed to methodology and theoretical guidance, supervised the project, managed administration, and edited the manuscript.

## Abbreviations

SCM: Structural Causal Models
DAG: Directed Acyclic Graph.
LLM: Large Language Model.
SBERT: Sentence-BERT embedding model.
MentalCLOUDS: Mental health Counseling cOmponent gUided Dialogue Summaries
5P: Framework of Predisposing, Precipitating, Perpetuating, Presenting, (and Protective) factors.